\documentclass{article}
\usepackage[hyperref]{emnlp2020}
\usepackage{times}
\usepackage{latexsym}

\usepackage{microtype}

\aclfinalcopy 
\def\aclpaperid{***} 

\usepackage{graphicx}
\usepackage{subfigure}
\usepackage{booktabs} 
\usepackage{tabularx}
\usepackage{amsmath}
\newcolumntype{i}{X}
\newcolumntype{j}{>{\hsize=.25\hsize}X}
\newcolumntype{k}{>{\hsize=.05\hsize}X}
\newcolumntype{J}{>{\hsize=.33\hsize}X}
\newcolumntype{K}{>{\hsize=.1\hsize}X}
\newcolumntype{R}{>{\raggedleft\arraybackslash}X}
\usepackage[T1]{fontenc}
\usepackage{amsfonts}
\usepackage{hyperref}
\usepackage{color}
\usepackage{multirow}
\usepackage{enumitem}
\definecolor{NavyBlue}{rgb}{0.0, 0.0, 0.5}

\title{Question Directed Graph Attention Network \\ for Numerical Reasoning over Text}

\author{Kunlong Chen$^{\dagger}$ \quad Weidi Xu$^{\dagger}$ \quad Xingyi Cheng\thanks{~~Corresponding author: derrickzy@gmail.com}~~$^{\dagger}$  \\ { \bf  Zou Xiaochuan$^{\dagger}$ \quad Yuyu Zhang$^{\mathsection}$ \quad Le Song$^{\dagger\mathsection}$ } \\ { \bf Taifeng Wang$^{\dagger}$ \quad Yuan Qi$^{\dagger}$  \quad Wei Chu$^{\dagger}$ } \\
$^{\dagger}$ Ant Group \\
{\tt  \{kunlong.ckl, weidi.xwd, fanyin.cxy, xiaochuan.zxc,} \\
{\tt
le.song, taifeng.wang, weichu.cw, yuan.qi\}@antgroup.com} \\
$^{\mathsection}$ College of Computing Georgia Institute of Technology \\
{\tt
\{yuyu\}@gatech.edu}
}

\date{}

\begin{document}

\maketitle

\begin{abstract}
Numerical reasoning over texts, such as addition, subtraction, sorting and counting, is a challenging machine reading comprehension task, since it requires both natural language understanding and arithmetic computation. 
To address this challenge, we propose a heterogeneous graph representation for the context of the passage and question needed for such reasoning, and design a question directed graph attention network to drive multi-step numerical reasoning over this context graph. 
%
%
Our model, which combines deep learning and graph reasoning, achieves remarkable results in benchmark datasets such as DROP~\footnote{{https://leaderboard.allenai.org/drop/submissions/public}. As of September 08, 2020, our models are ranked first in the case of fair comparison using the identical pre-training model. The code is available on https://github.com/emnlp2020qdgat/QDGAT}. 

\end{abstract}

\section{Introduction}

\setlength{\abovedisplayskip}{4pt}
\setlength{\abovedisplayshortskip}{1pt}
\setlength{\belowdisplayskip}{4pt}
\setlength{\belowdisplayshortskip}{1pt}



Machine reading comprehension (MRC) aims to develop AI models that can answer questions for text documents. 
Recently, the performance of MRC in public datasets has been improved dramatically due to the advanced pre-trained models, such as BERT~\cite{devlin-etal-2019-bert}, RoBERTa~\cite{DBLP:journals/corr/abs-1907-11692} and ALBERT~\cite{DBLP:journals/corr/abs-1909-11942}.

However, pre-trained models are not explicitly aware of the concepts of numerical reasoning since numeracy supervision signals are rarely available during pre-training. 
The representations from these pre-trained models fall short in their ability to support downstream numerical reasoning.  
Yet such ability is critical for the comprehension of financial news and scientific articles, since basic numerical operations, such as addition, subtraction, sorting and counting, need to be conducted to extract the essential information~\cite{DBLP:conf/naacl/DuaWDSS019}.

Recently, \citet{DBLP:conf/naacl/DuaWDSS019} proposed a numerically-aware QANet (NAQANet), which treats the span extractions, counting, and numerical addition/subtraction separately.
However, this work is preliminary in the sense that the model neglects the relative magnitude between numbers.
To improve this method, \citet{DBLP:conf/emnlp/RanLLZL19} proposed NumNet, which constructs a number comparison graph that encodes the relative magnitude information between numbers on directed edges.
Although NumNet achieves superior performance than other numerically-aware models~\cite{hu-etal-2019-multi,DBLP:conf/emnlp/AndorHLP19,ggb2020injecting,DBLP:conf/iclr/ChenLYZSL20}, we argue that NumNet is insufficient for sophisticated numerical reasoning, since it lacks two critical ingredients for numerical reasoning:
\begin{enumerate}[leftmargin=*,nolistsep,nosep]
\setlength\itemsep{0.5em}
    \item \textbf{Number Type and Entity Mention.} The number comparison graph in NumNet is not able to identify different number types, and lacks the information of entities mentioned in the document that connect the number nodes.
    \item \textbf{Direct Interaction with Question.} The graph reasoning module in NumNet leaves out the direct question representation, which may encounter difficulties in locating important numbers directed by the question as the pivot for numerical reasoning.
\end{enumerate}
    



\begin{table*}[t]
\scriptsize
\caption{Two MRC cases requiring numerical reasoning are illustrated. There are entities and numbers of different types. Both are emphasized by different colors: \textcolor{orange}{entity}, \textcolor{red}{number}, \textcolor{blue}{percentage}, \textcolor{NavyBlue}{date}, \textcolor{cyan}{ordinal}. We explicitly encode the type information into our model and leverage the question representation to conduct the reasoning process. 
}
\begin{tabularx}{\linewidth}{jik}
\toprule
Question & Passage & Answer \\ \midrule
At the battle of \textcolor{orange}{Caiboaté} how many \textcolor{orange}{Spanish} and \textcolor{orange}{Portuguese} were injured or killed? & 
... In \textcolor{NavyBlue}{1754} \textcolor{orange}{Spanish} and \textcolor{orange}{Portuguese} military forces were dispatched to force the \textcolor{orange}{Guarani} to leave the area ... Hostilities resumed in \textcolor{NavyBlue}{1756} when an army of \textcolor{red}{3,000} \textcolor{orange}{Spanish}, \textcolor{orange}{Portuguese}, and native auxiliary soldiers under \textcolor{orange}{José de Andonaegui} and \textcolor{orange}{Freire de Andrade} was sent to subdue the \textcolor{orange}{Guarani} rebels. On \textcolor{NavyBlue}{February 7, 1756} the leader of the \textcolor{orange}{Guarani} rebels, \textcolor{orange}{Sepé Tiaraju}, was killed in a skirmish with \textcolor{orange}{Spanish} and \textcolor{orange}{Portuguese} troops. ... \textcolor{red}{1,511} \textcolor{orange}{Guarani} were killed and \textcolor{red}{152} taken prisoner, while \textcolor{red}{4} \textcolor{orange}{Spanish} and \textcolor{orange}{Portuguese} were killed and about \textcolor{red}{30} were wounded...& 34 \\ \midrule
In which quarter did \textcolor{orange}{Stephen Gostkowski} kick his shortest field goal of the game? &
The \textcolor{orange}{Cardinals' east coast} struggles continued in \textcolor{cyan}{the second quarter} as quarterback \textcolor{orange}{Matt Cassel} completed a \textcolor{red}{15}-yard touchdown pass to running back \textcolor{orange}{Kevin Faulk} and an \textcolor{red}{11}-yard touchdown pass to wide receiver \textcolor{orange}{Wes Welker}, followed by kicker \textcolor{orange}{Stephen Gostkowski}'s \textcolor{red}{38}-yard field goal. In \textcolor{cyan}{the third quarter}, \textcolor{orange}{Arizona}'s deficit continued to climb as \textcolor{orange}{Cassel} completed a \textcolor{red}{76}-yard touchdown pass to wide receiver \textcolor{orange}{Randy Moss}, followed by \textcolor{orange}{Gostkowski}'s \textcolor{red}{35}- and \textcolor{red}{24}-yard field goal. In \textcolor{cyan}{the fourth quarter}, \textcolor{orange}{New England} concluded its domination with \textcolor{orange}{Gostkowski}'s \textcolor{red}{30}-yard & third  \\
\bottomrule
\end{tabularx}

\label{tab:example}
\end{table*}

\begin{figure*}
\includegraphics[width=\linewidth]{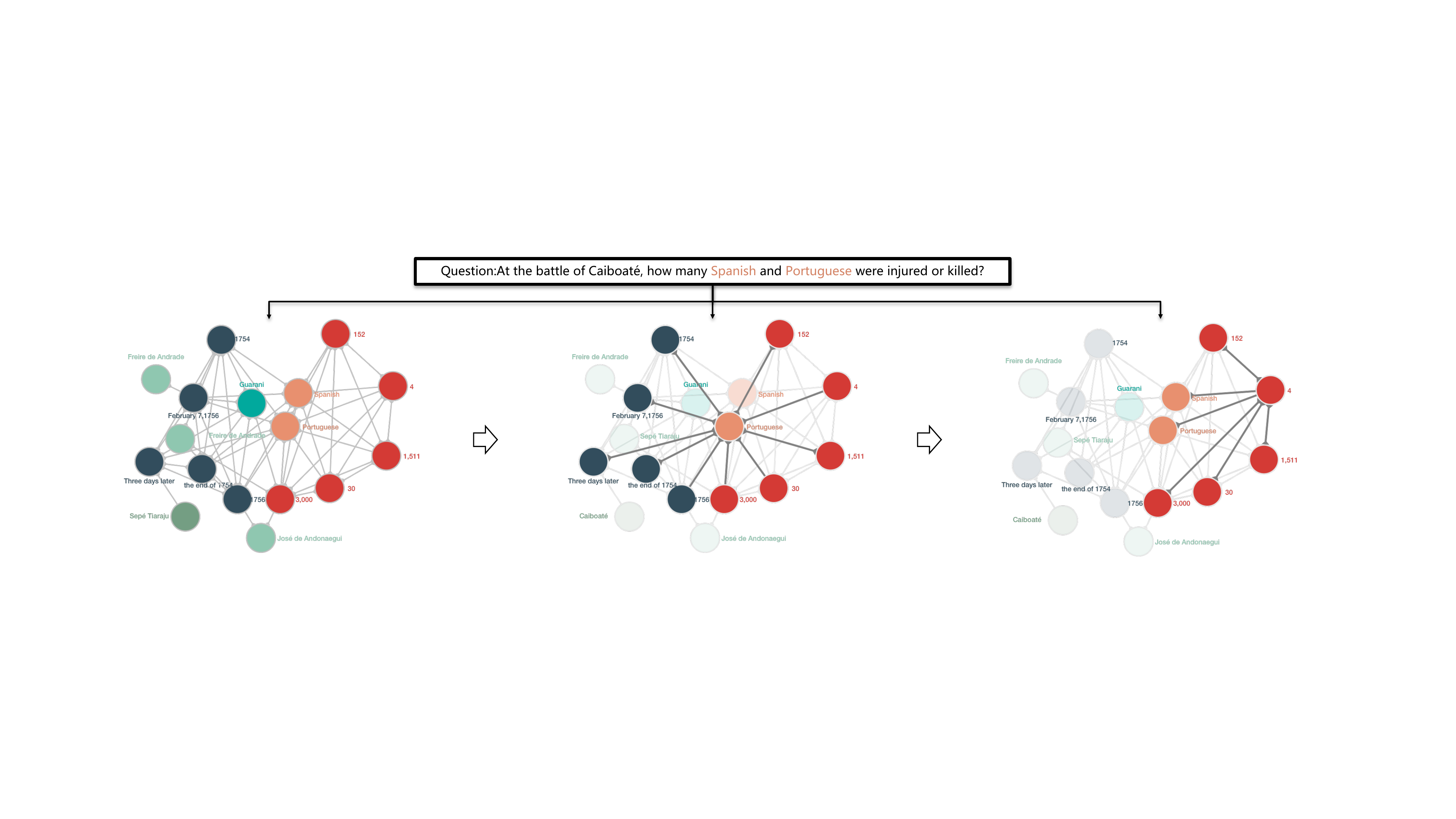}
\vspace{-5mm}
\caption{The constructed heterogeneous typed graph of the example in Table~\ref{tab:example} is illustrated on the left. 
The \textcolor{red}{red} (\textcolor{NavyBlue}{dark blue}) nodes are the numbers (dates) and the others are entities.
The edges encode the relations among the numbers and entities:
(1) The numbers with the same number type, e.g., date, are wired together.
(2) The graph connects the numbers and the entities that are in the same sentence to indicate their co-occurrence.
In the first round, the model pays attention to a sub-graph that contains the \textcolor{orange}{\emph{Spanish}} and \textcolor{orange}{\emph{Portuguese}} entities since they are mentioned in the question.
In the update, the model learns to distinguish between the numbers and the dates and extracts the numbers related to the question.
In the second round, the representations of the numbers are updated by the messages from the entities as well as the question to conduct the reasoning.
}
\label{fig:illustration}
\end{figure*}

The number type and entity information play essential roles in numerical comprehension and reasoning.
As per the study in the cognitive system - ``this abstract, notation-independent appreciation of numbers develops gradually over the first several years of life ... human infants appreciate numerical quantities at a non-symbolic level: They know approximately how many objects they see before them even though they do not understand number words or Arabic numerals.'', the concept of discrete number is gradually developed through the real-life experience~\cite{cantlon2009neural}.
The association among the numbers and entities is a strong regularization for learning the numerical reasoning model: the comparison and addition/subtraction between numbers are typically applied to those with the same type or referring to the same entity. 
To illustrate it, we show two concrete examples of numerical reasoning over texts in Table~\ref{tab:example}.
In the first example, a question related to the ``population'' is being asked. There are 5 ``\textit{people counting}'' numbers and 3 ``\textit{date}'' numbers.
When the type of number is given, the reasoning difficulty is largely reduced if the model learns to extract the ``\textit{people counting}'' numbers conditioned on this ``population'' question. 
In addition, the entities in the graph provide explicit information on the correlation between the passage and the question.
The entities in the question may occur in several sentences in the passage, indicating how each number is related to each other through these bridging entities, which helps the QA model better collect and aggregate the information for numerical reasoning. We also observe that when the question entities co-occur in a single sentence (the last sentence in this example), this could be a hint that the answer can be derived from that sentence.
The second example illustrates the case in span extraction.
Similarly, the model is benefited when the correlations between the numbers and ``Stephen Gostkowski'' are explicitly provided.



To explicitly integrate the type and entity information into the model, we construct a heterogeneous directed graph where the nodes consist of entities and different types of numbers, and the edges can encode different types of relations. 
The corresponding graph of the example in Table~\ref{tab:example} is illustrated in Figure~\ref{fig:illustration}.
The graph nodes are composed of entities and numbers from both the question and the passage.
The numbers of the same type are densely connected with each other. The co-occurred numbers and entities within a sentence are also connected with each other. 



Based on this heterogeneous graph, we propose a question directed graph attention network (QDGAT) for the task of numerical MRC.
As the answer-related numbers can be directed by the question, QDGAT incorporates the contextual encoding of the question in the graph reasoning process.
More specifically, QDGAT employs a contextual encoder, such as BERT~\cite{devlin-etal-2019-bert} and RoBERTa~\cite{DBLP:journals/corr/abs-1907-11692}, to extract the representations of the numbers and entities in both the question and the passage, serving as the initial embeddings of each node in the graph.
With the heterogeneous graph, QDGAT learns to collect information from the graph conditioned on the question for numerical reasoning.
Each node is also described by a context-aware representation conditioned on the question, and the representations are updated through a message-passing iteration.
After multiple iterations of message passing with graph neural networks, QDGAT gradually aggregates the node information to answer the question.
In this sense, QDGAT abstracts the representation of passage and question in a way more consistent with human perception and reasoning, making the model produces a more interpretable reasoning pattern.

We evaluate QDGAT on two benchmark datasets: the DROP dataset~\cite{DBLP:conf/naacl/DuaWDSS019} which requires Discrete Reasoning Over the content of Paragraph, and a subset of the RACE dataset~\cite{lai-etal-2017-race} that contains the number-related questions.
Experimental results indicate that QDGAT achieves remarkable performance on the DROP dataset, currently ranked as top 1 for all released models. And also rank first compared with other models that use the identical pre-training model.





\section{Related Work}

{\bf Machine Reading Comprehension.}
Benefit from recent improvements of pre-trained deep language models like BERT~\cite{devlin-etal-2019-bert}, XLNet~\cite{yang2019xlnet}, a considerable progress of MRC have been made on the annotated datasets such as SQuAD~\cite{rajpurkar-etal-2016-squad}, RACE~\cite{lai-etal-2017-race}, TriviaQA~\cite{JoshiTriviaQA2017} and so on. 
To answer complex questions of MRC, a number of neural architectures have been proposed such as Attentive Reader~\cite{hermann2015teaching}, BiDAF~\cite{Seo2017Bidirectional}, Gated Attention Reader~\cite{dhingra-etal-2017-gated}, R-NET~\cite{wang2017rnet}, QANet~\cite{wei2018qanet}, which achieved excellent results on existing datasets.
Some recent works~(LCGN~\cite{hu2019lcgn}, NMNs~\cite{gupta2020nms}, NumNet~\cite{DBLP:conf/emnlp/RanLLZL19}) attaching reasoning capabilities to models shows a promising direction. LCGN uses graph neural networks~(GNN) conditioned on the input questions to support rational reasoning. NMNs parse the questions into one of several programs, each of which is responsible for specific reasoning ability.

{\bf Numerical Reasoning in MRC.}
Numerical reasoning has been studied when solving arithmetic word problems~(AWP). However, existing AWP models only worked on small datasets, and the arithmetic expression must be clearly given. Numerical reasoning in MRC is more challenging since the numbers and reasoning rules are extracted from raw text, which requires a more sophisticated model.
NAQANet improved the output layer of QANet to predict the answers from the arithmetic computation over numbers. 
In addition to NAQANet, GenBERT~\cite{ggb2020injecting} injects numerical skills into BERT by generating numerical data.
~\cite{DBLP:conf/iclr/ChenLYZSL20} provides a semantic parser that points to locations in the text that can be used in further numerical operations.
BERT-Calculator~\cite{DBLP:conf/emnlp/AndorHLP19} defines a set of executable programs and learns to choose one to derive numerical answers. 
NumNet~\cite{DBLP:conf/emnlp/RanLLZL19} uses a numerically-aware graph neural network to encode numbers, which made further progress on the DROP dataset.
However, the graph in NumNet contains only numbers and ignores their types and context information which play a key point in numerical reasoning. 
Our model differs from NumNet in two aspects: (1) We use a heterogeneous graph containing entities and different types of numbers to encode the relations among the entities and numbers, rather than the relations from numerical comparison; (2) We use the question embedding to modulate the attention over graph neighbors and update the representation to achieve reasoning.

\begin{figure*}
\centering
\includegraphics[width=\textwidth]{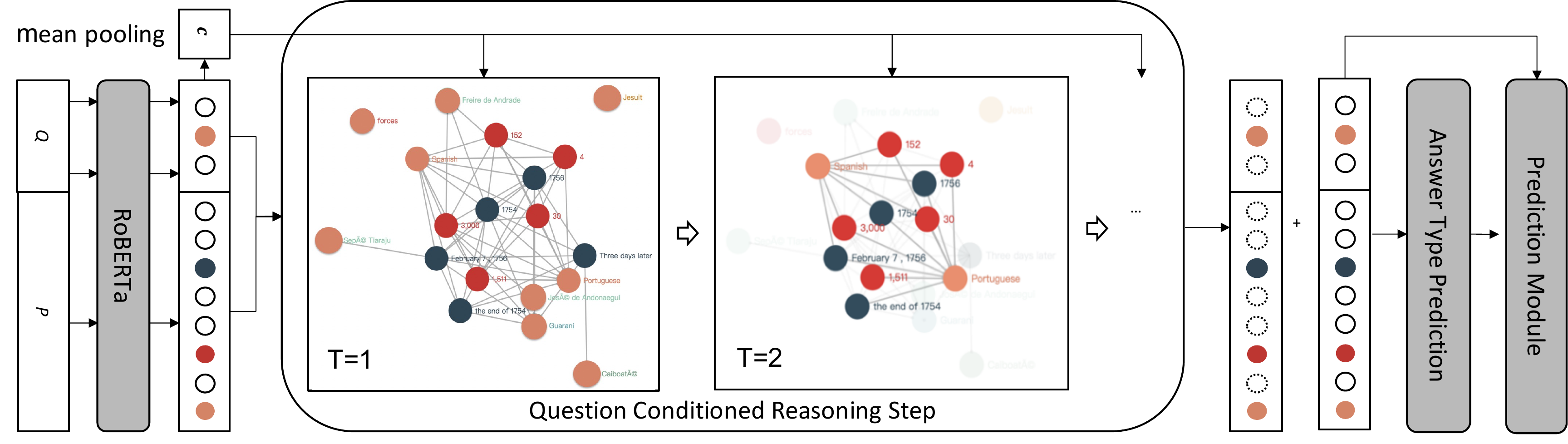}
\vspace{-4mm}
\caption{
The framework of our model. It consists of a representation extractor (left), a reasoning module (middle) and a prediction module (right).
The reasoning module reasons over a heterogeneous directed graph whose nodes are the numbers and the entities.
Two kinds of relations are encoded: (1) the numbers of the same type are connected with each other by the type-specific edges, (2) the entities and the numbers are connected when they co-occur in a sentence.
The reasoning is conditioned on the question explicitly to guide the message propagation over the graph.
In each iteration, each node selectively receives the messages from the neighboring nodes with the question representation to update its representation.
The derived representations of these nodes are then combined with the RoBERTa output for the final prediction module.
The dashed circle means zero vector.
} 
\label{fig:framework}
\end{figure*}

\section{Method}
In this section, we first introduce the machine reading comprehension task requiring numerical reasoning. Then the framework of our model is provided, followed by detailed descriptions about its components.

\subsection{Problem Definition}
In the MRC task, each data sample consists of a passage $P$ and a related question $Q$.
The goal of an MRC model is to answer the question according to $P$.
Besides predicting the text spans as in the standard MRC tasks, the answer $A$ in the case of numerical reasoning can also be a number derived from arithmetic computations, such as sorting, counting, addition and subtraction.



\subsection{Overall Framework}
The framework of the proposed model is briefly depicted in Figure~\ref{fig:framework}.
The model is composed of three main components, i.e., a representation extractor module, a reasoning module, and a prediction module.
The representation extractor is responsible for semantic comprehension.
Upon the extractor, a heterogeneous graph with typed numbers and related entities is constructed.
To aggregate the information between the numbers and entities, we propose a question directed graph attention network (QDGAT) to make sophisticated reasoning.
This graph attention network directly employs the question $Q$ to manage the message passing over the typed graph.

{\bf Word Representation Extractor.}
We employ RoBERTa \cite{DBLP:journals/corr/abs-1907-11692} as the base architecture for the representation of textual inputs.
The module takes the passage $P$ and the question $Q$ as input and outputs representation vectors for each token:
\begin{equation}
\hat{\mathbf{Q}}, \hat{\mathbf{P}} =  \mathtt{RoBERTa}({Q}, {P}) \,,
\end{equation}
where $\mathtt{RoBERTa}$ denotes the transformer encoder initialized with RoBERTa  parameters, 
$\hat{\mathbf{P}}$ ($\hat{\mathbf{Q}}$) denotes the list of the token vectors of size $d_h$ in the passage (question).
It takes the concatenation of \texttt{[CLS]}, ${Q}$, \texttt{[SEP]}, ${P}$ and \texttt{[SEP]} as input, and outputs representations of $\mathbf{Q}$ and $\mathbf{P}$ as $\hat{\mathbf{Q}}$ and $\hat{\mathbf{P}}$.  

{\bf Graph Construction.}
This module builds the heterogeneously typed graph from text data. The graph $\mathcal{G}=(\mathbf{V}, \mathbf{E})$ contains numbers $\mathbf{N}$ and entities $\mathbf{T}$ as the nodes $\mathbf{V}=\{\mathbf{N}, \mathbf{T}\}$, and its edges $\mathbf{E}$ encode the information of the number type and the relationship between the numbers and the entities. 
The details will be clarified in Section~\ref{sec:graph}.


{\bf Numerical Reasoning Module.}
The numerical reasoning module, i.e., QDGAT, is built upon the representation and graph extractor.
Based on the graph $\mathcal{G}=(\mathbf{V}, \mathbf{E})$, the QDGAT network can be formulated as follows:
\begin{align}
\mathbf{M}^Q &= \mathbf{W}^M\hat{\mathbf{Q}} \,, \\
\mathbf{M}^P &= \mathbf{W}^M\hat{\mathbf{P}} \,, \\
\mathbf{c} &= \mathbf{W}^c\texttt{MEAN}(\hat{\mathbf{Q}}) \,, \\
\mathbf{U} &= \texttt{QDGAT}(\mathcal{G}; \mathbf{M}^P, \mathbf{M}^Q, \mathbf{c}) \,, \label{equ:qdgat}
\end{align}
where $\mathbf{W}^M \in \mathbb{R}^{d_h\times d_h}$ is a shared projection matrix to obtain the input of QDGAT, \texttt{MEAN} denotes the mean pooling, $\mathbf{W}^c \in \mathbb{R}^{d_h\times d_h}$ projects the averaged vector of the representations in the question to derive $\mathbf{c}$.
$\mathbf{c}$ is the question language embedding used to direct the reasoning in \texttt{QDGAT}.
QDGAT then reasons over the representations ($\mathbf{M}^P$, $\mathbf{M}^Q$) and the graph $\mathcal{G}$ conditioned on the question command $\mathbf{c}$.

{\bf Prediction Module}
The prediction module takes the output of graph reasoning network $\mathbf{U}$ for final prediction.
At present, the types of answers are generally divided into three categories in NAQANet and NumNet+: (a) span extraction, (b) count, (c) arithmetic expression.
We implemented separate modules for these answer types and all of them take the output of graph network $\mathbf{U}$ and question embedding $\mathbf{c}$ as input.
They are specified as follows:

\begin{itemize}
\item Span extraction: There are three span extraction tasks, i.e., single passage span, multiple passage spans, single question span. The probability for single span extraction is derived by the product of the probabilities of the start and end positions in either question or passage. For multiple spans extraction, the probability is constructed referring to~\cite{DBLP:journals/corr/abs-1909-13375}.
\item Count: This problem is regarded as a 10-class classification problem (0-9), which covers about 97\% counting problems in the DROP dataset.
\item Arithmetic expression: The answer is derived by an arithmetic computation. In the DROP dataset, only addition and subtraction operations are involved. We achieved this by classifying each number into one of $(-1, 0, +1)$, which is then used as the coefficient of the number in the numerical expression to arrive at the final answer.
\end{itemize}
We used a unique classification network to classify the data sample into one of five fine-grained types ($T$).
And each type solver employs a unique output layer to calculate the conditional answer probability $p({A}|{T})$.


\subsection{Graph Construction with Typed Number and Entities}
\label{sec:graph}

Here, we illustrate how to construct the heterogeneous graph $\mathcal{G}=(\mathbf{V}, \mathbf{E})$ in our model.
NumNet solely concerns the numerical comparisons between numbers by using the directed edges.
The graph used in our model differs from NumNet significantly: Rather than modeling the numerical comparison, our graph instead exploits two sources of information, i.e., the type of numbers and the related entities.
As illustrated in Figure~\ref{fig:framework}, the nodes of graph $\mathbf{V}$ consists of both entities $\mathbf{T}$ and numbers $\mathbf{N}$, both of which are recognized by an external name entity recognition (NER) system~\footnote{We used Standford CoreNLP toolkit~\cite{DBLP:conf/acl/ManningSBFBM14}.}.

Specifically, the NER software labels each token in the text into one of 21 pre-defined categories.
The tokens labeled as \texttt{NUMBER}, \texttt{PERCENT}, \texttt{MONEY}, \texttt{TIME}, \texttt{DATE}, \texttt{DURATION}, \texttt{ORDINAL} are regarded as the numbers.
Since DROP dataset contains a lot of samples related to American football games, we also used heuristic rules to extract the numbers of \texttt{YARD} type in the data samples.
Besides, we leveraged a number extractor, i.e., word2num~\footnote{https://pypi.org/project/word2number/}, to extract the remaining numbers, which are labeled as \texttt{NUMBER}.
All these tokens construct the number set $\mathbf{N}$ with 8 number types ($\mathcal{V}_N = (\texttt{NUMBER}, \texttt{PERCENT}, \texttt{MONEY}, \texttt{TIME}, \texttt{DATE}, \\ \texttt{DURATION},  \texttt{ORDINAL}, \texttt{YARD}$)). 
As for other recognized tokens,  we map them into the label \texttt{ENTITY} to build the entity set $\mathbf{T}$ whose type set $\mathcal{V}_T$ is $\texttt{ENTITY}$.
In the following, we use $t(v)\in \mathcal{V}_N \cup \mathcal{V}_T$ to indicate the type of the node.
The type information can directly inform the model to find the numbers related to the question and thus reduces the reasoning difficulty.

The edges $\mathbf{E}$ encode the relationship among the numbers and the entities, which correspond to two situations.
\begin{itemize}[leftmargin=*,nolistsep,nosep]
\setlength\itemsep{0.5em}
\item \textit{The edge between the numbers}: An edge $e_{i,j}$ exists between two numbers $v_i$ and $v_j$ if and only if these two numbers are of the same type in $\mathcal{V}_N$. And its relation $r_{i,j}=r_{j,i}$ corresponds to the number type.
\item \textit{The edge between the entity and the number}: An edge $e_{i,j}$ exists between an entity $v_i$ and a number $v_j$ if and only if $v_i$ and $v_j$ co-occur in the same sentence. In this situation, the relation $r_{i,j}=r_{j,i}$ is $\texttt{ENT+DIGIT}$.
\end{itemize}
The edges in the first situation cluster the same typed numbers together, which provides an evident clue to help to reason over the numbers.
In the second situation, we assume that an entity is relevant to a number when they appear closely.
This kind of edges roughly indicates the correlations between the numbers and the entities in most cases. On the other hand, the relative magnitude relations in Numnet+ are not considered in our graph since early experiments with these relations did not improve results.
Overall, the graph has 9 relations $\mathcal{R}$, i.e., 8 relations for number types and 1 relation for $\texttt{ENT+DIGIT}$. 

\subsection{Question Directed Graph Attention Network}
Here, we present the details of the \texttt{QDGAT} function.
Based on the heterogeneous graph $\mathcal{G}$, our QDGAT makes context-aware numerical reasoning conditioned on the question, which collects the relational information through multiple iterations of message passing between the numbers and the entities.
It dynamically determines which objects to interact with through the edges in the graph, and sends messages through the graph to propagate the relational information.
To achieve this, we augment the reasoning module with the contextualized question representation.
For instance in the example in Table~\ref{tab:example}, the task is to find how many \emph{Spanish} and \emph{Portuguese} were injured or killed.
The entities and the numbers are explicitly marked and are modeled in a heterogeneous graph, as shown in Figure~\ref{fig:illustration}. 
Our model is able to extract the related entities, i.e., the \emph{Spanish} and \emph{Portuguese}, conditioned on $\mathbf{c}$.
Among the numbers related to these two entities, a number of them are of date type, while the others are about people.
However, only the numbers related to people should be concerned as requested by the question. 
Then the model reasons over these numbers to derive the expression for the answer calculation.

{\bf Module Input.}
The graph neural network takes the representations from the extractor as the input.
Each node is represented by the corresponding vector in $\mathbf{M}^P$ and $\mathbf{M}^Q$. 
Formally, when $v_i$ is in the passage, the input of node $v_i$ is the $\mathbf{v}_i = \mathbf{M}^{P}[\mathbf{I}^P(v_i)]$, where $\mathbf{I}^P$ returns the index of $v_i$ in $\mathbf{M}^P$~\footnote{When $v_i$ corresponds to several tokens, the average of these vectors is used.}.
The collected vectors from the question and the passage construct the input of reasoning module $\mathbf{v}^0$.

{\bf Question Directed Node Embedding Update.}
At each iteration $t \in \{1,...T\}$, a question directed layer integrates the question information with the current node embedding representations. This step is to mimic the reasoning step of detecting relevant nodes. More specifically, the question, represented by $\mathbf{c}$, is used to direct the information propagation between the nodes (i.e., the numbers and the entities).
Each node collects the information from the neighbors with the question command.
The role of numbers and entities is not only dependent on the input itself, but also the neighbors and the relations between them.
Therefore, we adopt the self-attention layer~\cite{vaswani2017attention} to dynamically aggregate the information.
The representation is first converted into three spaces denoting the query, key and value, conditioned on $\mathbf{c}$:
\begin{align}
\mathbf{m}^t &= \mathbf{W}_{dc}^t{g(\mathbf{W}_{fc} \mathbf{c})} \,, \\
\mathbf{x}_q^t &= \mathbf{W}_{qv}[\mathbf{v}^t: \mathbf{v}^0]\odot \mathbf{W}_{qc}\mathbf{m}^t\,, \\
\mathbf{x}_k^t &= \mathbf{W}_{kv}[\mathbf{v}^t: \mathbf{v}^0] \odot \mathbf{W}_{kc}\mathbf{m}^t\,, \\
\mathbf{x}_v^t &= \mathbf{W}_{vv}[\mathbf{v}^t: \mathbf{v}^0] \odot \mathbf{W}_{vc}\mathbf{m}^t\,,
\end{align}
where $\mathbf{m}^t$ denotes the command vector extracted dynamically from the $\mathbf{c}$ with $\mathbf{W}_{dc}^t$ and $\mathbf{W}_{fc} \in \mathtt{R}^{d_h \times 2d_h}$, $g$ denotes the ELU activation function~\cite{DBLP:journals/corr/ClevertUH15},
$\mathbf{W}_{qv}$, $\mathbf{W}_{kv}$ and $\mathbf{W}_{vv}$ are of size $d_h \times 2d_h$, $\mathbf{W}_{qc}$, $\mathbf{W}_{kc}$ and $\mathbf{W}_{vc}$ are of size $d_h \times d_h$, $[a:b]$ means the concatenation of $a$ and $b$, and $\odot$ means the element-wise multiplication.
These equations include the input $\mathbf{v}^0$ to maintain the original information.

{\bf Directed Graph Attention.} At each iteration, this graph attention layer for each node aggregates information from the neighbors of the node. This step is to mimic the reasoning step of selecting the relevant relations to operate on. 
More specifically, we compute the relatedness between the node $i$ and $j$, which is measured by summarizing all relations:
\begin{align}
a^t_{i,j} = f(\sum_{ r\in \mathcal{R}_{i,j}} \mathbf{W}^r_{a} [{\mathbf{x}_q^t}_{,i}: {\mathbf{x}_k^t}_{,j}]) \,,
\end{align}
where $\mathcal{R}_{i,j}$ means the relations between the two nodes, $a_{i,j}$ denotes the attention score of the node $i$ for the node $j$, $\mathbf{W}^k_a$ is the vector to map the representations into a scalar for the relation $r$ and $f$ denotes the leakyReLU activation function~\cite{DBLP:journals/corr/XuWCL15}. 

This attention score is used in the message propagation to collect the right amount of information from each neighboring node.
In the propagation function, the calculation of the node interaction is as follows:
\begin{align}
\alpha^t_{i,j} &= \frac{\exp (a^t_{i,j})}{\sum_{j' \in \mathcal{N}_i} \exp (a^t_{i,j'})} \,, \\
\hat{\mathbf{x}}^t_i &= \sum_{j \in \mathcal{N}_i} \alpha_{i,j} \mathbf{x}_{v,j} \,,\\
\mathbf{v}^{t+1}_i &= \mathbf{W}_u [\mathbf{v}^t_i ; \hat{\mathbf{x}}^t_i]  \,,
\end{align}
where $\mathcal{N}_i$ contains the adjacent nodes of the node $i$ in the $\mathcal{G}$ and $\mathbf{W}_u$ is in  $\mathbb{R}^{d_h\times 2d_h}$.
With the weight $\alpha_{i,j}$ obtained, the values of neighboring nodes are summarized to derive a new representation $\hat{\mathbf{x}}$.
Finally, the new representation of $\mathbf{v}$ is computed by mapping the concatenation of $\mathbf{v}^0$ and $\hat{\mathbf{x}}$.

We denote the node embedding update and the graph attention layers as a function:
\begin{equation}
\mathbf{v}^{t+1} = \texttt{QDGAT-single}(\mathcal{G}, \mathbf{v}^t, \mathbf{c}) \,.
\end{equation}
From the process of this reasoning step, we can see that the module receives the information from the question, which directly manages the message propagation among the numbers and the entities. 

\paragraph{Module Output}
We perform $T$ iterations of the reasoning step of $\texttt{QDGAT-single}$ to perform $\texttt{QDGAT}$ in Equation~\ref{equ:qdgat}.
The output of the last layer $\mathbf{v}^T$ is obtained for the numbers and entities in $\mathbf{U}$.
For other tokens, the representation vectors from the extractor are used.
Formally, the calculation of the output $\mathbf{U}$ is implemented as follows:
\begin{equation}
\mathbf{U}_i = 
\begin{cases}
    \mathbf{M}_i + \mathbf{v}^T_{\mathbf{J}(i)},& \text{if } i\text{-th token} \in \mathbf{V} \\
    \mathbf{M}_i, & \text{otherwise } 
\end{cases}
\end{equation}
where $\mathbf{J}(i)$ denotes the index of token $i$ in the graph nodes, $\mathbf{M}$ denotes the combination of $\mathbf{M}^P$ and $\mathbf{M}^Q$ for simplicity.
$\mathbf{U}$ is then used in the prediction module for the five answer types mentioned above. 

\section{Experiments}

\subsection{Dataset and Evaluation Metrics}

We performed experiments on the DROP dataset~\cite{DBLP:conf/naacl/DuaWDSS019}, which was recently released for research on numerical machine reading comprehension (MRC). DROP is constructed by crowd-sourcing question-answer pairs on passages from Wikipedia, which contains 77,409 / 9,536 / 9,622 samples in the original training / development / testing split.
Following the previous work~\cite{DBLP:conf/naacl/DuaWDSS019}, we used Exact Match (EM) and F1 score as the evaluation metrics.

\subsection{Baselines}
We choose publicly available methods (including non-published ones on the dataset leaderboard) as our baselines:
\begin{itemize}[leftmargin=*,nolistsep,nosep]
\setlength\itemsep{0.5em}
    \item \textbf{Semantic parsing models:} Syn Dep, OpenIE and SRL~\cite{DBLP:conf/naacl/DuaWDSS019}. All these models are enhanced versions of KDG~\cite{krishnamurthy-etal-2017-kdg} with different sentence representations.
    \item \textbf{Traditional MRC models:} (1) BiDAF, a model that uses a bi-directional attention flow network to obtain a query-aware context representation; (2) QANet, a model that combines convolution and self-attention models to answer the questions; (3) BERT~\cite{devlin-etal-2019-bert}, a pre-trained deep Transformer~\cite{vaswani2017attention} model that has improved results on many NLP tasks.
    \item \textbf{MRC models with numerical reasoning module:} (1) NAQANet~\cite{DBLP:conf/naacl/DuaWDSS019}, a model that adapts the output layer of QANet to numeric reasoning; (2) ALBERT-Calculator~\cite{DBLP:conf/emnlp/AndorHLP19}, a model based on ALBERT-xxlarge~\cite{DBLP:conf/iclr/LanCGGSS20} that picks one of executable programs from a predefined set to derive numerical answers. (3) NumNet, a model that embeds numerical properties into the distributed representation by using a GNN on the number graph; (4) NumNet+~\footnote{https://github.com/llamazing/numnet\_plus}, an enhanced version of NumNet, which uses a pre-trained RoBERTa model and supports multi-span answers. 
\end{itemize} 

\subsection{Experiment Settings}
We use the large RoBERTa model as the contextual encoder, with 24 layers, 16 attention heads, and 1024 embedding dimensions.
This indicates that the hidden size $d_h$ is 1024.
The model was trained end-to-end for 5 epochs using Adam optimizer~\cite{DBLP:journals/corr/KingmaB14} with a batch size of 16.
For the hyperparameters of RoBERTa, the learning rate is 5e-5 and the L2 weight decay is 1e-6.
For the other parts, the learning rate is 1e-4 and the L2 weight decay is 5e-5.
We perform $T=4$ iterations of the graph reasoning step, which performs best in our experiments.
We adopt the standard data pre-processing following previous work~\cite{DBLP:conf/emnlp/RanLLZL19}.

\subsection{Main Results}
The overall experimental results are reported in Table~\ref{overall-results-drop-table}, where the performance of baseline methods is obtained from previous work~\cite{DBLP:conf/naacl/DuaWDSS019,Seo2017Bidirectional,DBLP:conf/emnlp/RanLLZL19,DBLP:conf/emnlp/AndorHLP19} and
the public leaderboard.\footnote{https://leaderboard.allenai.org/drop/submissions/public}

The first three methods in Table~\ref{overall-results-drop-table} are based on either semantic parsing or information extraction, and perform poorly on the numerical MRC task.
Traditional MRC methods BiDAF and QANet, which has no numerical reasoning modules, achieve slightly better performance but are still far from satisfying.
Methods that are customized for numerical reasoning, including NAQANet and NumNet, have achieved significantly better performance in terms of EM and F1 score. Compared to traditional MRC methods, these methods can handle different answer types, e.g., span extraction, counting, and addition/subtraction of numbers.


Our method QDGAT outperforms all the existing methods, achieving 86.38 F1 score and 83.23 EM on the test set, which narrows the human performance gap to less than 11 points.
NumNet+ is the most relevant one to our method, which also leverages a graph neural network as well as the RoBERTa contextual encoder. Compared to NumNet+, QDGAT incorporates the number types and entity mentions into the graph attention network, and directs the graph reasoning process with the question. In this way, our method can better capture the relations between numbers and entities, and also reduce the learning difficulty due to the interaction with the question during the graph reasoning.
Experimental results demonstrate the effectiveness of QDGAT, which outperforms NumNet+ by 1.23 in terms of EM and 1.37 in terms of F1 score.
Ensembling three of our models with different random seeds and learning rates further improves the performance.

\begin{table}[t]
\caption{Overall results on the development and test set of DROP. For QDGAT$_{p}$, we used more careful data pre-processing and a RoBERTa pre-trained on the SQuaD dataset. $\dagger$ denotes that the result is taken from the public leaderboard. Better results are in bold.}
\label{overall-results-drop-table}
\vskip 0.15in
\begin{center}
\begin{small}
\begin{tabularx}{\linewidth}{lRRRRRRRr}
\toprule
\multirow{2}{*}{Method} & \multicolumn{2}{c}{Dev}  & \multicolumn{2}{c}{Test} \\ \cmidrule(lr){2-3} \cmidrule(lr){4-5}
 & EM  & F1 & EM & F1  \\
\midrule
Syn Dep & 9.38 & 11.64 & 8.51 & 10.84\\
OpenIE & 8.80 & 11.31 & 8.53 & 10.77 \\
SRL & 9.28 & 11.72 & 8.98 & 11.45 \\
BiDAF & 26.06 & 28.85 & 24.75 & 27.49 \\
QANet & 27.50 & 30.44 & 25.50 & 28.36 \\
BERT & 30.10 & 33.36 & 29.45 & 32.70 \\
NAQANet & 46.20 & 49.24 & 44.07 & 47.01 \\
ALBERT-Calculator & 80.22 & 83.98 & 79.85 & 83.56 \\
NumNet & 64.92 & 68.31 & 64.56 & 67.97 \\
NumNet+~(RoBERTa) & 81.07$^\dagger$ & 84.42$^\dagger$ & 81.52$^\dagger$ & 84.84$^\dagger$ \\ 
NumNet+~(ensemble) & 82.63$^\dagger$ & 85.59$^\dagger$ &83.14$^\dagger$& 86.16$^\dagger$ \\ \midrule
QDGAT~(RoBERTa) & \textbf{82.74} & \textbf{85.85} & \textbf{83.23} & \textbf{86.38} \\
QDGAT$_{p}$~(RoBERTa) & \textbf{84.07} & \textbf{87.05} & \textbf{84.53} & \textbf{87.57} \\
QDGAT$_{p}$~(ensemble) & \textbf{85.31} & \textbf{88.10} & \textbf{85.46} & \textbf{88.38} \\ \midrule 
Human & & & 94.09 & 96.42 \\
\bottomrule
\end{tabularx}
\end{small}
\end{center}
\vskip -0.1in
\end{table}

\subsection{Ablation Analysis}
To examine the impact of different components of QDGAT, we conduct ablation studies and compare the performance in Table~\ref{setting-results-drop-table}. 
QDGAT$_{\rm NH}$ removes the number type and entity from the graph, and QDGAT$_{\rm NQ}$ removes question direction from QDGAT and instead uses a normal graph convolution message passing mechanism. 
NumNet+ serves as a baseline for reference, since it has no question attention, no entities and no number types in the graph.
We observe that QDGAT$_{\rm NQ}$, which has no question directed attention, performs worse. This justifies that the reasoning with graph neural network is more effective when conditioned on the input question.
We also observe that QDGAT$_{\rm NH}$ performs significantly worse, which demonstrates the importance of incorporating the information of number types and entity mentions in the reasoning graph. 
This is consistent with our intuition that numbers with the same type or connected to the same entity are more relevant to each other. 

\begin{table}[t]
\caption{Ablation study results on the development set of DROP. QDGAT$_{\rm NH}$ removes the number type and entity from the graph, and QDGAT$_{\rm NQ}$ removes question direction from QDGAT. Better results are in bold.}
\label{setting-results-drop-table}
\vskip 0.15in
\begin{center}
\begin{small}
\begin{tabularx}{\linewidth}{lRRRr}
\toprule
Method & EM & F1  \\
\midrule
NumNet+ & 81.07 & 84.42  \\
QDGAT$_{\rm NH}$ & 81.98 & 84.94 \\
QDGAT$_{\rm NQ}$ & 82.04 & 85.01  \\
QDGAT & \textbf{82.74} & \textbf{85.85}  \\
\bottomrule
\end{tabularx}
\end{small}
\end{center}
\vskip -0.1in
\end{table}

Table~\ref{category-results-drop-table} decomposes the QA performance on different answer types in the development set of DROP. 
As reported in the table, QDGAT works better on the questions relating to numbers and dates, which requires more specific numerical reasoning compared with the span extraction.
The remarkable improvement indicates that the proposed method effectively benefits the reasoning module to comprehend the numerical problems.
Notably, the performance in span extraction can still be improved by our method.
The span extraction in DROP heavily relies on the ability to comprehend the relation between the number and the entity (c.f. the second example in Table~\ref{tab:example}).

\begin{table}[t]
\caption{Decomposed performance on different answer types in the development set of DROP. Better results are in bold.}
\label{category-results-drop-table}
\vskip 0.12in
\begin{center}
\begin{scriptsize}
\begin{tabularx}{\linewidth}{lrrrrrrrr}
\toprule
\multirow{2}{*}{Method} & \multicolumn{2}{c}{Number}  & \multicolumn{2}{c}{Date} & \multicolumn{2}{c}{Span} \\ \cmidrule(lr){2-3} \cmidrule(lr){4-5} \cmidrule(lr){6-7}
       & EM  & F1 & EM & F1 & EM & F1  \\
\midrule
NumNet+ & 82.89 & 83.13 & 56.67 & 63.91 & 82.00 & 86.84 \\
QDGAT & \textbf{86.00} & \textbf{86.23} & \textbf{60.27} & \textbf{67.48} & \textbf{84.05} & \textbf{88.53}\\
\bottomrule
\end{tabularx}
\end{scriptsize}
\end{center}
\vskip -0.1in
\end{table}


\begin{table*}[!h]
\caption{The cases from the DROP dataset. The predictions from the QDGAT and NumNet+ are illustrated. The differences between the output of these two models demonstrate the properties of the proposed model. The last two columns indicate the arithmetic expression, obtained by assigning a sign~(plus, minus or zero) for each extracted numbers~(we omitted the zero sign numbers). Then the answer was derived by summing up the signed numbers.}
\label{examples-drop-table}
\vskip 0.15in
\begin{center}
\scriptsize
\begin{tabularx}{\linewidth}{JiKK}
\toprule
Question \& Answer & Passage & NumNet+ & QDGAT \\ \midrule
\textbf{Q:} How many less in  age \textcolor{blue}{percentage} in teenagers than adult? \newline
\textbf{A:} 1.3 & The age distribution, in Aigle is; 933 children or \textcolor{blue}{10.7\%} of the population are between \textcolor{red}{0} and \textcolor{red}{9} years old and 1,137 teenagers or \textcolor{blue}{13.0\%} are between \textcolor{red}{10} and \textcolor{red}{19}. Of the adult population, 1,255 people or \textcolor{blue}{14.3\%} of the population are between 20 and 29 years old... & 19-13.0-10=-4 & 14.3-13.0\newline=1.3 \\ \midrule
\textbf{Q:} How many \textcolor{red}{yards} did \textcolor{orange}{Kasay} kick? \newline
\newline
\textbf{A:} 94 & ... Carolina scored first in the second quarter with kicker \textcolor{orange}{John Kasay} hitting a \textcolor{red}{45-yard} field goal . The Falcons took the lead with QB Joey Harrington completing a \textcolor{red}{69-yard} TD pass to WR Roddy White . The Panthers followed up with QB Jake Delhomme completing a \textcolor{red}{13-yard} TD pass to RB DeShaun Foster ... In the fourth quarter , the Panthers scored again , with \textcolor{orange}{Kasay kicking} a \textcolor{red}{49-yard} field goal . The Falcons ' Andersen nailed a \textcolor{red}{25-yard} field goal to end the scoring ...   & 
+45=45 &45+49\newline=94 \\ \midrule
\textbf{Q:} How many \textcolor{blue}{months} after \textcolor{orange}{Mengistu Haile Mariam} was made head of state did \textcolor{orange}{Ethiopia} close the U.S. military mission and the communications centre?
 \newline
\textbf{A:} 2 & ... A sign that order had been restored among the Derg was the announcement of \textcolor{orange}{Mengistu Haile Mariam} as head of state on \textcolor{blue}{02/1977}. However, the country remained in chaos as the military attempted to suppress its civilian opponents in a period known as the Red Terror ... \textcolor{orange}{Ethiopia} closed the U.S. military mission and the communications centre in \textcolor{blue}{04/1977}. In \textcolor{blue}{06/1977}, \textcolor{orange}{Mengistu} accused Somalia of infiltrating SNA soldiers into the Somali area to fight alongside the WSLF. Despite considerable evidence to the contrary... & Count: 3 &+4-2=2 \\
\bottomrule
\end{tabularx}
\end{center}
\label{tab:cases}
\end{table*}


\subsection{Performance on RACENum}
To investigate the generalization capability of QDGAT in numerical reasoning, we examine whether the pre-trained model on DROP is transferable.
We compare QDGAT with NumNet+ on RACE~\cite{lai-etal-2017-race}, a dataset collected from the English exams for middle and high school Chinese students.  
We extracted a special part of examples from RACE, where the questions start with ``how many'', referred to as RACENum.
RACENum is then divided into middle school exam~(RACENum-M) and high school exam~(RACENum-H) categories.
The RACENum-M and RACENum-H datasets contain 633 and 611 questions accordingly.
Since the original RACE dataset is in the multiple-choice form, we converted them into the DROP data format.
The accuracy of NumNet+, QDGAT and its ablation variants on RACENum are summarized in Table~\ref{race-results-race-table}, which is consistent with the performance comparison on the DROP dataset.

The overall low scores are attributed to the lack of training on the in-domain data.
QDGAT achieves 43.7 points on RACENum on average, which is approximately 4.5 points higher than NumNet+.
Both QDGAT$_{\rm NQ}$ and QDGAT$_{\rm NH}$ still outperform NumNet+ by a 2--3 points margin.
We further confirmed that ablating either the entity information or question attention from the heterogeneous graph weakens the power of QDGAT to learn numeracy and the capability of understanding numbers in either digits or word form. 
Compared with QDGAT, ablating the question directed attention, i.e., QDGAT$_{\rm NQ}$, leads to about a 1 point drop. For QDGAT$_{\rm NH}$ that removes the number type and entity mentions from the graph, it performs consistently worse than QDGAT, demonstrating the impact of the heterogeneous graph for numerical reasoning. 


\begin{table} 
\caption{The accuracy on the unsupervised RACENum dataset.}
\label{race-results-race-table}
\vskip 0.15in
\begin{center}
\begin{small}
\begin{tabularx}{\linewidth}{lRRRRr}
\toprule
Method & RACE-M & RACE-H  & Avg. \\
\midrule
NumNet+ & 46.98 & 31.59 & 39.29  \\
QDGAT$_{\rm NH}$  & 50.88 & 35.30 & 43.09 \\
QDGAT$_{\rm NQ}$ & 49.67 & \textbf{35.84} & 42.76 \\
QDGAT & \textbf{52.53} & 34.86 & \textbf{43.70} \\
\bottomrule
\end{tabularx}
\end{small}
\end{center}
\vskip -0.1in
\end{table}

\subsection{Case Study}
We show several examples to provide insights into how our model works. Table~\ref{examples-drop-table} compares the different model prediction results from NumNet+ and QDGAT:
\begin{itemize}
\setlength\itemsep{0.5em}
\item The first example shows the importance of number types. NumNet+ treats all numbers as the same type, which fails to capture that the question only cares about percentage and incorrectly predicts ``\textit{19}'' (type age) as part of the result. In contrast, QDGAT extracts the relevant numbers and derives the correct answer.
\item The second example highlights the importance of entity mentions. NumNet+ fails to extract ``\textit{49-yard}'', but QDGAT easily captures this number since ``\textit{49-yard}'' and ``\textit{45-yard}'' are connected to the same entity ``\textit{Kassy}'' on the heterogeneous graph which is generated from the passage.
\item The third example shows the importance of question conditioning. 
Solving this example requires to extract the two dates related to two events mentioned in the question.
Without direct interaction between the question, the model tends to recognize this example as a counting problem since the question starts with ``how many''.
However, when combined with question directed attention, correct numbers can be filtered out.
\end{itemize}

\section{Conclusion}
In this work, we propose a novel method named QDGAT for numerical reasoning in the machine reading comprehension task.
Our method not only builds a more compact graph containing different types of numbers, entities, and relations, which can be a general method for other sophisticated reasoning tasks but also conditions the reasoning directly on the question language embedding, which modulates the attention over graph neighbors and change messages being passed iteratively to achieve reasoning.
The experimental results verify the effectiveness of our method. 
In the future, we plan to extend our model to learn the heterogeneous graph automatically, which assures more flexibility for numerical reasoning. We would also explore to learn the types of numbers and entities together the reasoning modules using variational autoencoder techniques~\cite{DBLP:journals/corr/KingmaW13}, which may help the NER system better adapt to the numerical reasoning task.




\bibliography{emnlp2020}
\bibliographystyle{acl_natbib}

\end{document}